\documentclass[a4paper, 10pt, conference]{ieeeconf}      
\usepackage{FG2024}
\usepackage{booktabs}
\usepackage{multirow}
\usepackage{graphicx}
\usepackage{subcaption}
\FGfinalcopy 

\IEEEoverridecommandlockouts                              
\usepackage{graphics} 
\usepackage{epsfig} 
\usepackage{mathptmx} 
\usepackage{times} 
\usepackage{amsmath} 
\usepackage{amssymb}  

\def\FGPaperID{1} 

\title{\LARGE \bf ENTIRe-ID: An Extensive and Diverse Dataset for Person Re-Identification}

\author{\parbox{16cm}{\centering
    {\large Serdar Yıldız$^{1,3}$ and Ahmet Nezih Kasım$^{2,3}$}\\
    {\normalsize
    $^1$ Department of Computer Engineering, Yıldız Technical University, Istanbul, Turkey\\
    $^2$ Department of Computer Engineering, Bogazici University, Istanbul, Turkey \\
    $^3$ BİLGEM, TÜBİTAK, Kocaeli, Turkey}}
}

\usepackage{fancyhdr}
\begin{document}

\ifFGfinal
\thispagestyle{empty}
\pagestyle{empty}
\else
\author{Anonymous FG2024 submission\\ Paper ID \FGPaperID \\}
\pagestyle{plain}
\fi
\maketitle

\thispagestyle{fancy}

\begin{abstract}

The growing importance of person re-identification in computer vision has highlighted the need for more extensive and diverse datasets. In response, we introduce the ENTIRe-ID dataset, an extensive collection comprising over 4.45 million images from 37 different cameras in varied environments. This dataset is uniquely designed to tackle the challenges of domain variability and model generalization, areas where existing datasets for person re-identification have fallen short. The ENTIRe-ID dataset stands out for its coverage of a wide array of real-world scenarios, encompassing various lighting conditions, angles of view, and diverse human activities. This design ensures a realistic and robust training platform for ReID models. The ENTIRe-ID dataset is publicly available at https://serdaryildiz.github.io/ENTIRe-ID

\end{abstract}

\section{INTRODUCTION}

The person re-identification (ReID) problem holds significant importance within the rapidly evolving domain of computer vision. It involves identifying and matching individuals across diverse scenarios and camera viewpoints, unlike face recognition systems that operate in controlled environments \cite{Gheissari}. Person ReID is crucial for various applications such as urban surveillance, public safety, retail analytics, and crowd management \cite{leng2019survey, martini2020open}. It enables tracking individuals across locations, providing advantages to law enforcement, retail security, and smart city initiatives, ultimately improving public safety and operational efficiency.

Person ReID, with its ability to recognize individuals beyond the limitations of facial features, stands as a complementary and, in some cases, superior solution to face recognition systems. The reliance on holistic person information encompassing clothing, body posture, and accessories, enables more robust and reliable identification, especially in scenarios where facial features are partially or fully obscured \cite{zheng2015partial}. This broader scope enhances the applicability of person ReID in diverse real-world environments.

However, the performance of person ReID systems is not without its challenges. Domain shift arising from variations in lighting, camera specifications, and environmental conditions poses a significant limitation, leading to performance degradation in real-world settings \cite{zheng2015scalable, li2014deepreid, wei2018person}. Moreover, the absence of large and diverse datasets magnifies these challenges, hindering the ability of models to generalize effectively across different scenarios.

Addressing the limitations of person ReID, we recognize the critical need for a large dataset that encapsulates the richness and diversity of real-world scenarios. A dataset of substantial scale and diversity becomes essential for training robust models that are capable of handling the intricacies of domain shift and ensuring reliable person matching across varying conditions.

\begin{figure}[t]
 \centering
 \includegraphics[width=0.67\columnwidth]{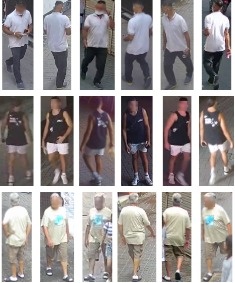}
 \caption{Examples of images taken at various times, angles, and cameras}
 \label{Fig:dataset-cam-exp}
\end{figure}

In this article, we present the ENTIRe-ID dataset, collected from 4 continents, with high diversity and the largest number of individuals and images in the person re-identification literature. Figure \ref{Fig:dataset-cam-exp} shows representative scenes obtained from the ENTIRe-ID dataset. The dataset significantly contributes to the ReID community in multiple dimensions. First and foremost, it sets a new standard for dataset size with an unmatched scale, incorporating 4.45 million images and 13,540 person IDs. Beyond sheer quantity, diversity is a fundamental aspect of our dataset, as it encompasses cameras from four continents, capturing a range of environmental conditions. The inclusion of real-world actions, such as carrying items, controlling vehicles, and engaging in everyday activities, enhances the dataset's authenticity. Furthermore, the incorporation of feature vectors compared to other datasets solidifies our contribution to the field's diversity, showcasing the dataset's expansive scope.

The paper continues with the following sections: Section \ref{sec:meth} provides the adopted methodology in the data collection process. Section \ref{sec:exp-res} presents the experimental results. Section \ref{sec:priv} gives information about privacy concerns. The paper is concluded in Section \ref{sec:conc}.

\begin{figure*}[]
 \centering
 \includegraphics[width=\textwidth]{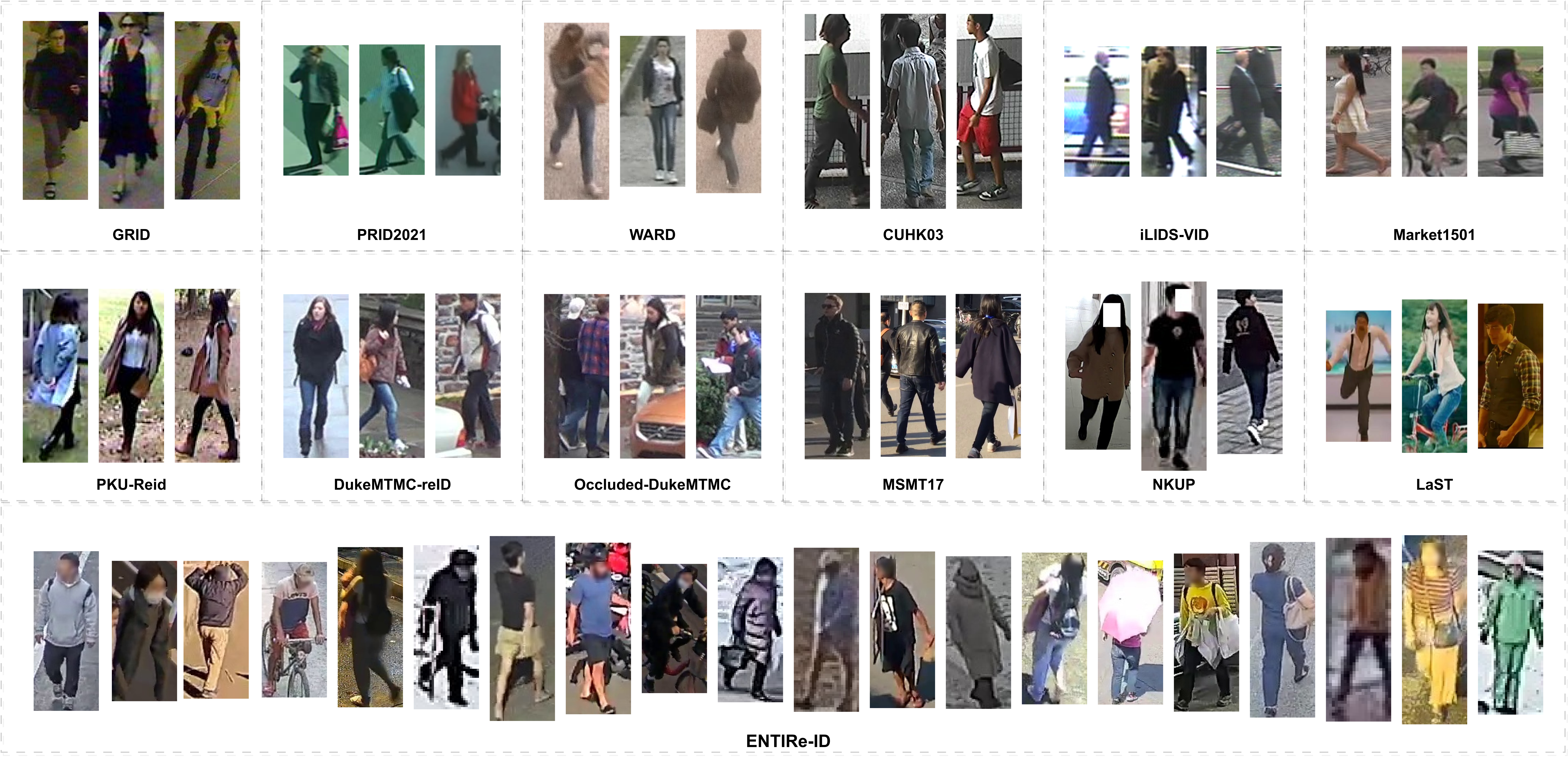}
 \caption{Samples of the ENTIRe-ID Dataset and Literature}
 \label{Fig:dataset-samples}
\end{figure*}

\begin{table}[t]
\caption{Person ReID Dataset Literature Review}
\label{table:comparison}
\begin{tabular}{@{}ccccc@{}}
\toprule
\textbf{Dataset Name} & \textbf{Year} & \textbf{Num. ID} & \textbf{Num. Cam} & \textbf{Num. Img} \\ \midrule
VIPer \cite{VIPer}                 & 2007          & 632              & 2                 & 1264                 \\
ETHZ \cite{ETHZ}                  & 2009          & 146              & 1                 & 8580                 \\
GRID \cite{GRID}                  & 2009          & 1025             & 8                 & 1275                 \\
CAVIAR4REID \cite{CAVIAR4REID}         & 2011          & 72               & 2                 & 1120                 \\
PRID2011 \cite{PRID2011}              & 2011          & 934              & 2                 & 24541                \\
V47 \cite{V47}                  & 2011          & 47               & 2                 & 752                  \\
WARD  \cite{WARD}                 & 2012          & 70               & 3                 & 4786                 \\
SAIVT-Softbio  \cite{SAIVT-Softbio}        & 2012          & 152              & 8                 & 64472                \\
CUHK01  \cite{CUHK01}               & 2012          & 971              & 2                 & 3884                 \\
CUHK02  \cite{CUHK02}               & 2013          & 1816             & 10                & 7264                 \\
CUHK03  \cite{CUHK03}               & 2014          & 1467             & 10                & 13164                \\
RAiD  \cite{RAiD}                 & 2014          & 43               & 4                 & 6920                 \\
HDA+  \cite{HDA+}                 & 2014          & 85               & 18                & 64028                \\
iLIDS-VID  \cite{iLIDS-VID}            & 2014          & 300              & 2                 & 42460                \\
PartialREID  \cite{PartialREID}          & 2015          & 60               & 2                 & 600                  \\
Market1501  \cite{Market1501}           & 2015          & 1501             & 6                 & 32217                \\
PKU-Reid  \cite{PKU-Reid}             & 2016          & 114              & 2                 & 1824                 \\
PRW  \cite{PRW}                  & 2016          & 932              & 6                 & 34304                \\
LSPS  \cite{LSPS}                 & 2016          & 8432             & -                 & 99809                \\
MARS  \cite{MARS}                 & 2016          & 1261             & 6                 & 1191003              \\
DukeMTMC-reID  \cite{DukeMTMC-reID}        & 2017          & 1812             & 8                 & 36441                \\
DukeMTMC4ReID  \cite{DukeMTMC4ReID}        & 2017          & 1852             & 8                 & 46261                \\
Airport  \cite{Airport}              & 2017          & 9651             & 6                 & 39902                \\
MSMT17  \cite{MSMT17}               & 2018          & 4101             & 15                & 126441               \\
RPIfield  \cite{RPIfield}             & 2018          & 112              & 12                & 601581               \\
NKUP  \cite{NKUP}                 & 2020          & 107              & 15                & 9738                 \\
LaST  \cite{LaST}                 & 2022          & 10862            & -                 & 228156               \\
NKUP+  \cite{NKUP+}                & 2022          & 361              & 29                & 40217                \\
DeepChange  \cite{DeepChange}           & 2023          & 1121             & 17                & 178407               \\ \midrule
\textbf{ENTIRe-ID}         & \textbf{2024} & \textbf{13540}       & \textbf{37}        & \textbf{4.45M}           \\ \bottomrule
\end{tabular}
\end{table}

\section{METHODOLOGY \label{sec:meth}}

Existing datasets are often limited in scope, collected with a few cameras in similar environments, which leads to images with similar visual patterns and noise characteristics. While models trained on these datasets perform well within their specific test environments, they often underperform when applied to more diverse and varied datasets. This constraint blocks the practical usability of even state-of-the-art models in real-life situations.

Although there have been major efforts to create ReID models that can be applied to many scenarios using domain adaptation, semi-supervised learning, and synthetic datasets, there is still a notable lack in the form of an extensive real-world person ReID dataset. To address this gap, we introduce the ENTIRe-ID dataset, comprising 4.45 million images and 13,540 unique person IDs. By collecting data from 37 publicly available Internet cams, this dataset is now the largest person ReID dataset based on our knowledge.

In the context of person ReID, Table \ref{table:comparison} systematically compiles the most notable datasets in the literature from 2007 to 2024. Particularly remarkable for its comprehensive scope, the ENTIRe-ID dataset surpasses other datasets in terms of the number of IDs, cameras, and images.

The LaST \cite{LaST} dataset, sharing a comparable number of IDs with the ENTIRe-ID dataset, diverges significantly in its data collection approach as it relies on movie data. Consequently, the scope of this dataset differs notably from other datasets described in the existing literature. Similarly, the MARS \cite{MARS} dataset, featuring a similar number of images, comprises less than 10\% of the number of IDs. Additionally, the NKUP+ \cite{NKUP+} dataset, closely aligned with the number of cameras, incorporates less than 1\% of the IDs found in the ENTIRe-ID dataset.

In Person ReID, a key step involves extracting feature vectors from cropped images of people, typically defined within bounding boxes. For Person ReID applications, obtaining cropped images is a prerequisite, usually achieved using the outputs from an object detection model. To overcome the challenges and potential biases of manual labeling, which is impractical for large datasets, we used an object detection model to automatically crop images of people. This approach not only ensures scalability but also minimizes human bias in the dataset creation process.

In this study, the YOLOv8 \cite{yolov8} model was chosen as the object detection model due to its widespread adoption, computational efficiency, and satisfactory performance. Individuals' crops that exceeded the specified thresholds for confidence score and pixel size were selected and included in the process of creating the data collection.

We adopted a deliberate frame sampling strategy to enhance computational efficiency and account for the low variability in successive video frames. This method, characterized by a significantly reduced sample rate in comparison to the FPS of the stream, aimed to capture a condensed yet representative set of 20–30 images per person. To ensure the accurate identification of the people detected in consecutive frames, it was essential to employ an object-tracking method. The ByteTrack \cite{ByteTrack} algorithm was chosen for this matching operation.

However, considering the possible difficulties in object tracking algorithms when faced with situations like frame dropping or failure of the object detection model to localize a person, an effective approach was implemented. Our tracking process included all frames from the video stream, even those containing objects with confidence scores below our set threshold. This inclusion not only ensures a robust tracking mechanism but also acts as a safeguard against errors inherent in both object detection and tracking models, thereby contributing to the overall reliability of the system.

After object detection and tracking processes, person sequences are obtained for each camera. Each sequence can contain a maximum of 250 images. The composition of these sequences is performed manually, both intra-camera and inter-camera. Errors such as ID switches and false detection are minimized manually. During the merging process, we compiled images of a single person (ID) taken from all cameras within a specific environment.

We partitioned the dataset into a training set with 10,799 IDs and a test set with 2,741 IDs. To enhance the accuracy of evaluating the generalization capabilities of person ReID models, all cameras in two environments were completely allocated for testing purposes. Furthermore, we enriched the test dataset with hand-selected samples of challenging cases from the cameras used in the training dataset. The test dataset consists of a total of 10415 gallery images and 3000 query images, which are hand-selected images from the sequences.

\begin{figure*}[t]
     \begin{subfigure}[t]{0.48\textwidth}
         \centering
         \includegraphics[width=\textwidth]{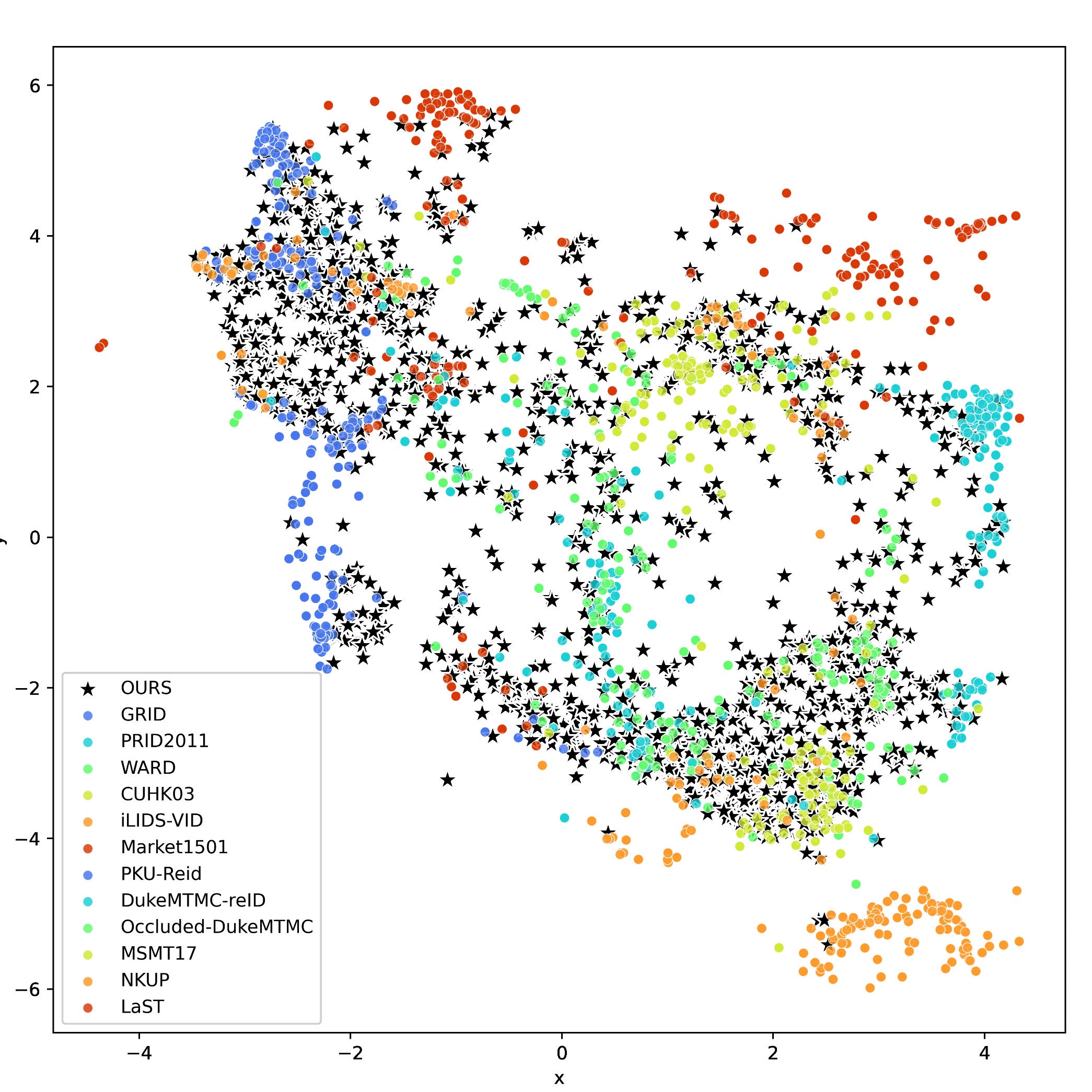}
         \caption{CLIP Feature Vector Visualization}
     \end{subfigure}     
     \begin{subfigure}[t]{0.48\textwidth}
         \centering
         \includegraphics[width=\textwidth]{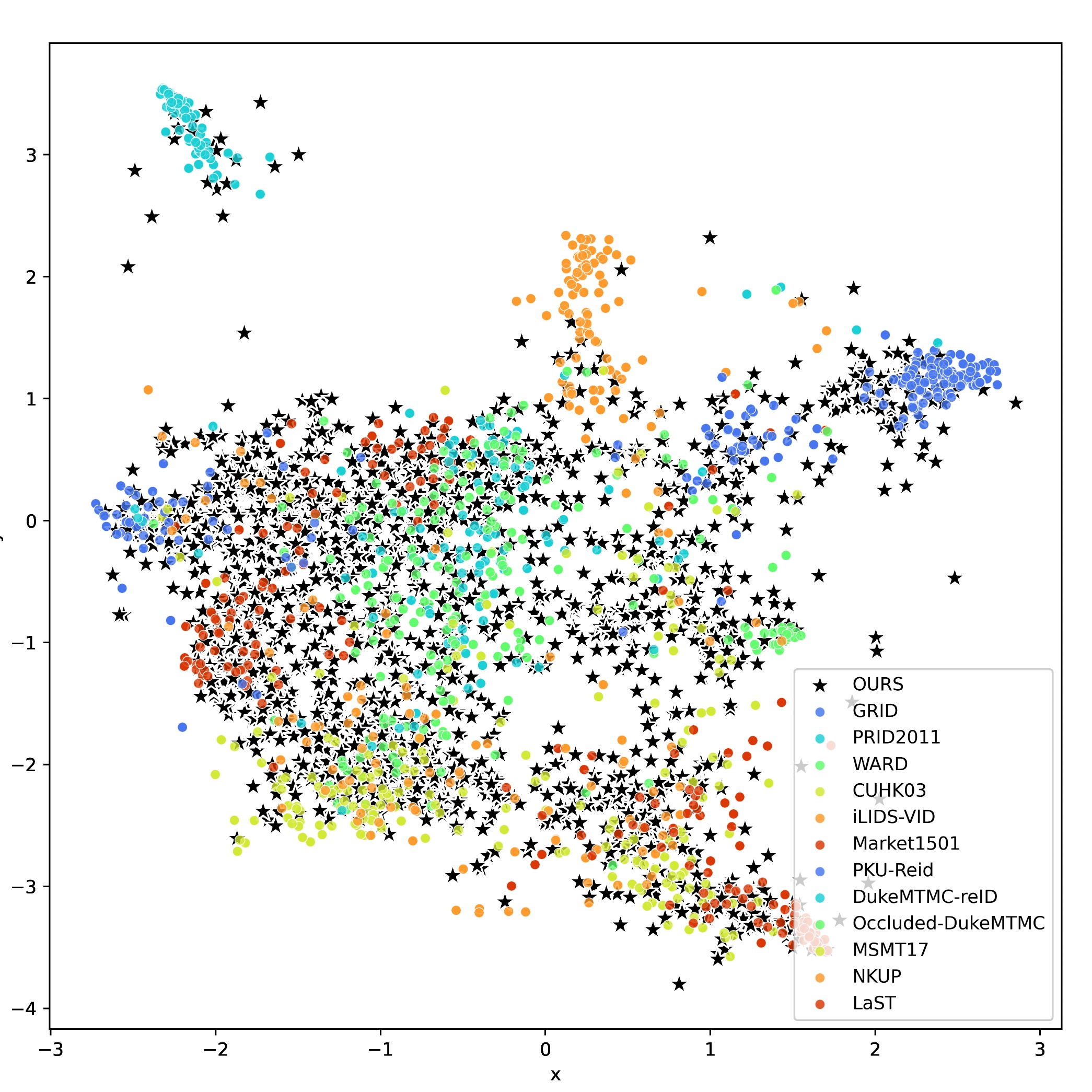}
         \caption{ImageNet Feature Vector Visualization}
     \end{subfigure}     
     \caption{t-SNE Visualization of Feature Vector Distributions}
     \label{Fig:tsne-viz}
\end{figure*}

\begin{table*}[t]
\centering
\caption{Performance Comparison of ReID Datasets}
\label{table:cross-test}
\begin{tabular}{@{}cccccccccccccccc@{}}
\toprule
\multirow{2}{*}{\textbf{\begin{tabular}[c]{@{}c@{}}Train\\ Dataset\end{tabular}}} & \multicolumn{3}{c}{\textbf{Market}}      & \multicolumn{3}{c}{\textbf{MSMT}}        & \multicolumn{3}{c}{\textbf{DukeMTMC}}    & \multicolumn{3}{c}{\textbf{Occ-Duke}} & \multicolumn{3}{c}{\textbf{ENTIRe-ID}}        \\
                                                                                  & \textbf{mAP} & \textbf{R1} & \textbf{R5} & \textbf{mAP} & \textbf{R1} & \textbf{R5} & \textbf{mAP} & \textbf{R1} & \textbf{R5} & \textbf{mAP}  & \textbf{R1}  & \textbf{R5} & \textbf{mAP} & \textbf{R1} & \textbf{R5} \\ \midrule
\textbf{Market}                                                                   & 87.2         & 94.7        & 98.2        & 12.8         & 32.3        & 44.7        & 41.6         & 58.6        & 72.4        & 14.8          & 19.2         & 31.6        & 27.0         & 26.9        & 41.1        \\
\textbf{MSMT}                                                                     & 39.5         & 66.2        & 79.4        & 61.8         & 81.9        & 90.6        & 49.9         & 66.7        & 80.4        & 22.9          & 29.7         & 46.4        & 38.4         & 38.0        & 56.3        \\
\textbf{DukeMTMC}                                                                 & 36.3         & 62.6        & 77.3        & 13.1         & 33.2        & 46.4        & 79.5         & 88.9        & 95.3        & 59.6          & 69.4         & 83.3        & 32.9         & 32.9        & 49.7        \\
\textbf{Occ-Duke}                                                            & 37.3         & 63.7        & 78.5        & 13.8         & 34.5        & 47.9        & 78.2         & 87.6        & 94.6        & 53.8          & 61.0         & 78.1        & 32.7         & 33.2        & 48.7        \\  \bottomrule
\end{tabular}
\end{table*}

\begin{table}[]
\centering
\caption{Impact of Facial Blurring on Model Performance}
\label{table:privacy}
\begin{tabular}{ccccccc}
\hline
\multirow{2}{*}{\textbf{\begin{tabular}[c]{@{}c@{}}Train\\ Dataset\end{tabular}}} & \multicolumn{3}{c}{\textbf{ENTIRe-ID Blured}}        & \multicolumn{3}{c}{\textbf{ENTIRe-ID}}  \\
                                                                                  & \textbf{mAP} & \textbf{R1} & \textbf{R5} & \textbf{mAP} & \textbf{R1} & \textbf{R5} \\ \hline
\textbf{Market}                                                                   & 27.0         & 26.9        & 41.1        & 27.6         & 27.0        & 42.0        \\
\textbf{MSMT}                                                                     & 38.4         & 38.0        & 56.3        & 40.5         & 39.7        & 58.9        \\
\textbf{DukeMTMC}                                                                 & 32.9         & 32.9        & 49.7        & 34.9         & 35.1        & 51.5        \\
\textbf{Occ-Duke}                                                            & 32.7         & 33.2        & 48.7        & 34.3         & 34.2        & 50.6        \\ \hline
\end{tabular}
\end{table}

The dataset consisted of IDs with images from a minimum of two cameras, according to research employing contrastive learning. Due to the absence of image elimination in the collected sequences, nearly all of the person crops in the dataset are included in the training set. The reason why elimination is not performed is to allow a model that has been pretrained inter-camera to be fine-tuned intra-camera. We believe this approach will create models with higher generalization performance. There are IDs in the training dataset that came from a minimum of 2 cameras and a maximum of 5 cameras. We focused on collecting numerous photos of the same individual from different cameras, as this approach improves the model's ability to generalize.

While it is important to have a large number of cameras, it is also important for the camera environments to be diverse. Besides the image quality is influenced by weather conditions like sunlight, snow, rain, and fog, cultural and seasonal factors affect the appearance of people in the images. To achieve diversity, we included cameras from four different continents in our dataset. As can be seen in the samples given in Figure \ref{Fig:dataset-samples}, all conditions that will increase diversity, such as rain, snow, sunny weather, night, and day are included in the dataset. In this way, differences in both camera quality and the appearance of the people were achieved.

Irrespective of the environment, some factors impact a person's identification feature vectors. The primary actions involve carrying an item and controlling a vehicle. Our dataset features individuals engaged in various activities, including riding bicycles or mopeds, carrying umbrellas or bags, pushing strollers, dragging trash cans, and walking pets on leashes.

\section{EXPERIMENTAL RESULTS \label{sec:exp-res}}

The scarcity of comprehensive datasets in Person ReID research has primarily led researchers to focus on improving model architectures. However, current state-of-the-art models are often trained on a limited range of datasets, leading to person ReID models that may exhibit increased epistemic uncertainty, significant biases, and reduced effectiveness in real-world scenarios.

To address these biases, we suggest comparing test results using well-known datasets like Market-1501 \cite{Market1501}, MSMT17 \cite{MSMT17}, DukeMTMC \cite{DukeMTMC-reID}, and Occluded-Duke \cite{DukeMTMC4ReID} alongside the ENTIRe-ID dataset. In this context, we present results obtained using the strong baseline model, solely based on the vision transformers implementation from the He et al. \cite{transreid} study, as shown in Table \ref{table:cross-test}. While each dataset yields the best results within its own training set, these models underperform when applied to datasets they weren't trained on. These results show how different the domains of the datasets are from each other. On examining the results from the ENTIRe-ID dataset, we find that the performance metrics are relatively consistent across different tests. Moreover, despite not being included in the training sets, the ENTIRe-ID dataset does not produce the lowest results in any of our experiments. This clearly indicates that the ENTIRe-ID dataset encompasses a broader range of scenarios compared to other datasets.

We evaluated the diversity of the ENTIRe-ID dataset compared to others by extracting feature vectors from publicly available datasets using the CLIP \cite{CLIP} and ImageNet ResNet \cite{resnet} models. These feature vectors were then projected in a two-dimensional space using t-SNE. Figure \ref{Fig:tsne-viz} illustrates the distribution of feature vectors. The divergence between the datasets is seen in the two feature vector distributions. As a result, the ENTIRe-ID dataset has distribution across nearly the full range. The LaST \cite{LaST} dataset, which exclusively consists of movie data, is not included because of its distinct movie domain in CLIP feature space. Furthermore, based on our analysis, the ENTIRe-ID dataset emerges as the most comprehensive dataset currently available in the literature.

\section{PRIVACY CONCERNS \label{sec:priv}}

Given that face recognition is not always feasible, person ReID serves as a complementary approach in such scenarios. Therefore, we focused on collecting a substantial number of low-resolution images to address this issue. However, most of the images in the dataset still clearly reveal the identities of the individuals depicted. We intentionally blurred the facial features in the images to prevent ReID models from recognizing and learning facial characteristics. More importantly, this action was primarily adopted to protect the privacy and rights of the individuals included in the dataset. The performance difference between blurred test images and unblurred test images is given in Table \ref{table:privacy}. Although a huge difference is not observed, the 1\%–2\% difference in mAP value can be interpreted as the model learning facial features and/or the model's uncertainty being high.

\section{CONCLUSIONS \label{sec:conc}}

The ENTIRe-ID dataset marks a major advancement in person ReID research, standing out with its unprecedented scale of 4.45 million images and 13,540 unique identities, gathered from 37 varied camera environments. This dataset uniquely tackles the challenges of domain shift and model generalization, making it highly applicable in real-world scenarios such as surveillance and other fields. The dataset's ethical design, which includes privacy-preserving facial blurring techniques, establishes it as a crucial public dataset for furthering person ReID research. We are confident that the unique characteristics of the ENTIRe-ID dataset will inspire and enable researchers to make significant strides in the field of person re-identification.



{\small
\bibliographystyle{ieee}
\bibliography{egbib}

\begin{thebibliography}{10}\itemsep=-1pt

\bibitem{SAIVT-Softbio}
A.~Bialkowski, S.~Denman, S.~Sridharan, C.~Fookes, and P.~Lucey.
\newblock A database for person re-identification in multi-camera surveillance networks.
\newblock In {\em 2012 International Conference on Digital Image Computing Techniques and Applications (DICTA)}, pages 1--8, 2012.

\bibitem{CAVIAR4REID}
D.~S. Cheng, M.~Cristani, M.~Stoppa, L.~Bazzani, V.~Murino, et~al.
\newblock Custom pictorial structures for re-identification.
\newblock In {\em Bmvc}, volume~1, page~6. Citeseer, 2011.

\bibitem{RAiD}
A.~Das, A.~Chakraborty, and A.~K. Roy-Chowdhury.
\newblock Consistent re-identification in a camera network.
\newblock In {\em Computer Vision--ECCV 2014: 13th European Conference, Zurich, Switzerland, September 6-12, 2014, Proceedings, Part II 13}, pages 330--345. Springer, 2014.

\bibitem{HDA+}
D.~Figueira, M.~Taiana, A.~Nambiar, J.~Nascimento, and A.~Bernardino.
\newblock The hda+ data set for research on fully automated re-identification systems.
\newblock In {\em Computer Vision-ECCV 2014 Workshops: Zurich, Switzerland, September 6-7 and 12, 2014, Proceedings, Part III 13}, pages 241--255. Springer, 2015.

\bibitem{Gheissari}
N.~Gheissari, T.~B. Sebastian, and R.~Hartley.
\newblock Person reidentification using spatiotemporal appearance.
\newblock In {\em 2006 IEEE computer society conference on computer vision and pattern recognition (CVPR'06)}, volume~2, pages 1528--1535. IEEE, 2006.

\bibitem{DukeMTMC4ReID}
M.~Gou, S.~Karanam, W.~Liu, O.~Camps, and R.~J. Radke.
\newblock Dukemtmc4reid: A large-scale multi-camera person re-identification dataset.
\newblock In {\em 2017 IEEE Conference on Computer Vision and Pattern Recognition Workshops (CVPRW)}, pages 1425--1434, 2017.

\bibitem{Airport}
M.~Gou, Z.~Wu, A.~Rates-Borras, O.~Camps, R.~J. Radke, et~al.
\newblock A systematic evaluation and benchmark for person re-identification: Features, metrics, and datasets.
\newblock {\em IEEE transactions on pattern analysis and machine intelligence}, 41(3):523--536, 2018.

\bibitem{VIPer}
D.~Gray and H.~Tao.
\newblock Viewpoint invariant pedestrian recognition with an ensemble of localized features.
\newblock In {\em Computer Vision--ECCV 2008: 10th European Conference on Computer Vision, Marseille, France, October 12-18, 2008, Proceedings, Part I 10}, pages 262--275. Springer, 2008.

\bibitem{resnet}
K.~He, X.~Zhang, S.~Ren, and J.~Sun.
\newblock Deep residual learning for image recognition.
\newblock In {\em Proceedings of the IEEE conference on computer vision and pattern recognition}, pages 770--778, 2016.

\bibitem{transreid}
S.~He, H.~Luo, P.~Wang, F.~Wang, H.~Li, and W.~Jiang.
\newblock Transreid: Transformer-based object re-identification.
\newblock In {\em Proceedings of the IEEE/CVF international conference on computer vision}, pages 15013--15022, 2021.

\bibitem{PRID2011}
M.~Hirzer, C.~Beleznai, P.~M. Roth, and H.~Bischof.
\newblock Person re-identification by descriptive and discriminative classification.
\newblock In {\em Image Analysis: 17th Scandinavian Conference, SCIA 2011, Ystad, Sweden, May 2011. Proceedings 17}, pages 91--102. Springer, 2011.

\bibitem{yolov8}
G.~Jocher, A.~Chaurasia, and J.~Qiu.
\newblock {Ultralytics YOLO}, Jan. 2023.

\bibitem{leng2019survey}
Q.~Leng, M.~Ye, and Q.~Tian.
\newblock A survey of open-world person re-identification.
\newblock {\em IEEE Transactions on Circuits and Systems for Video Technology}, 30(4):1092--1108, 2019.

\bibitem{iLIDS-VID}
M.~Li, X.~Zhu, and S.~Gong.
\newblock Unsupervised tracklet person re-identification.
\newblock {\em IEEE transactions on pattern analysis and machine intelligence}, 42(7):1770--1782, 2019.

\bibitem{CUHK02}
W.~Li and X.~Wang.
\newblock Locally aligned feature transforms across views.
\newblock In {\em Proceedings of the IEEE conference on computer vision and pattern recognition}, pages 3594--3601, 2013.

\bibitem{CUHK01}
W.~Li, R.~Zhao, and X.~Wang.
\newblock Human reidentification with transferred metric learning.
\newblock In K.~M. Lee, Y.~Matsushita, J.~M. Rehg, and Z.~Hu, editors, {\em Computer Vision -- ACCV 2012}, pages 31--44, Berlin, Heidelberg, 2013. Springer Berlin Heidelberg.

\bibitem{li2014deepreid}
W.~Li, R.~Zhao, T.~Xiao, and X.~Wang.
\newblock Deepreid: Deep filter pairing neural network for person re-identification.
\newblock In {\em Proceedings of the IEEE conference on computer vision and pattern recognition}, pages 152--159, 2014.

\bibitem{CUHK03}
W.~Li, R.~Zhao, T.~Xiao, and X.~Wang.
\newblock Deepreid: Deep filter pairing neural network for person re-identification.
\newblock In {\em 2014 IEEE Conference on Computer Vision and Pattern Recognition}, pages 152--159, 2014.

\bibitem{NKUP+}
M.~Liu, Z.~Ma, T.~Li, Y.~Jiang, and K.~Wang.
\newblock Long-term person re-identification with dramatic appearance change: Algorithm and benchmark.
\newblock In {\em Proceedings of the 30th ACM International Conference on Multimedia}, pages 6406--6415, 2022.

\bibitem{GRID}
C.~C. Loy, T.~Xiang, and S.~Gong.
\newblock Multi-camera activity correlation analysis.
\newblock In {\em 2009 IEEE Conference on Computer Vision and Pattern Recognition}, pages 1988--1995. IEEE, 2009.

\bibitem{PKU-Reid}
L.~Ma, H.~Liu, L.~Hu, C.~Wang, and Q.~Sun.
\newblock Orientation driven bag of appearances for person re-identification.
\newblock {\em arXiv preprint arXiv:1605.02464}, 2016.

\bibitem{WARD}
N.~Martinel and C.~Micheloni.
\newblock Re-identify people in wide area camera network.
\newblock In {\em 2012 IEEE Computer Society Conference on Computer Vision and Pattern Recognition Workshops}, pages 31--36, 2012.

\bibitem{martini2020open}
M.~Martini, M.~Paolanti, and E.~Frontoni.
\newblock Open-world person re-identification with rgbd camera in top-view configuration for retail applications.
\newblock {\em IEEE Access}, 8:67756--67765, 2020.

\bibitem{CLIP}
A.~Radford, J.~W. Kim, C.~Hallacy, A.~Ramesh, G.~Goh, S.~Agarwal, G.~Sastry, A.~Askell, P.~Mishkin, J.~Clark, et~al.
\newblock Learning transferable visual models from natural language supervision.
\newblock In {\em International conference on machine learning}, pages 8748--8763. PMLR, 2021.

\bibitem{DukeMTMC-reID}
E.~Ristani, F.~Solera, R.~Zou, R.~Cucchiara, and C.~Tomasi.
\newblock Performance measures and a data set for multi-target, multi-camera tracking.
\newblock In {\em European conference on computer vision}, pages 17--35. Springer, 2016.

\bibitem{ETHZ}
W.~R. Schwartz and L.~S. Davis.
\newblock Learning discriminative appearance-based models using partial least squares.
\newblock In {\em 2009 XXII Brazilian Symposium on Computer Graphics and Image Processing}, pages 322--329, 2009.

\bibitem{LaST}
X.~Shu, X.~Wang, X.~Zang, S.~Zhang, Y.~Chen, G.~Li, and Q.~Tian.
\newblock Large-scale spatio-temporal person re-identification: Algorithms and benchmark.
\newblock {\em IEEE Transactions on Circuits and Systems for Video Technology}, 32(7):4390--4403, 2022.

\bibitem{NKUP}
K.~Wang, Z.~Ma, S.~Chen, J.~Yang, K.~Zhou, and T.~Li.
\newblock A benchmark for clothes variation in person re-identification.
\newblock {\em International Journal of Intelligent Systems}, 35(12):1881--1898, 2020.

\bibitem{V47}
S.~Wang, M.~Lewandowski, J.~Annesley, and J.~Orwell.
\newblock Re-identification of pedestrians with variable occlusion and scale.
\newblock In {\em 2011 IEEE International Conference on Computer Vision Workshops (ICCV Workshops)}, pages 1876--1882, 2011.

\bibitem{wei2018person}
L.~Wei, S.~Zhang, W.~Gao, and Q.~Tian.
\newblock Person transfer gan to bridge domain gap for person re-identification.
\newblock In {\em Proceedings of the IEEE conference on computer vision and pattern recognition}, pages 79--88, 2018.

\bibitem{MSMT17}
L.~Wei, S.~Zhang, W.~Gao, and Q.~Tian.
\newblock Person transfer gan to bridge domain gap for person re-identification.
\newblock In {\em Proceedings of the IEEE conference on computer vision and pattern recognition}, pages 79--88, 2018.

\bibitem{LSPS}
T.~Xiao, S.~Li, B.~Wang, L.~Lin, and X.~Wang.
\newblock End-to-end deep learning for person search.
\newblock {\em arXiv preprint arXiv:1604.01850}, 2(2):4, 2016.

\bibitem{DeepChange}
P.~Xu and X.~Zhu.
\newblock Deepchange: A long-term person re-identification benchmark with clothes change.
\newblock In {\em Proceedings of the IEEE international conference on computer vision (ICCV)}, 2023.

\bibitem{ByteTrack}
Y.~Zhang, P.~Sun, Y.~Jiang, D.~Yu, F.~Weng, Z.~Yuan, P.~Luo, W.~Liu, and X.~Wang.
\newblock Bytetrack: Multi-object tracking by associating every detection box.
\newblock In {\em European Conference on Computer Vision}, pages 1--21. Springer, 2022.

\bibitem{MARS}
L.~Zheng, Z.~Bie, Y.~Sun, J.~Wang, C.~Su, S.~Wang, and Q.~Tian.
\newblock Mars: A video benchmark for large-scale person re-identification.
\newblock In {\em Computer Vision--ECCV 2016: 14th European Conference, Amsterdam, The Netherlands, October 11-14, 2016, Proceedings, Part VI 14}, pages 868--884. Springer, 2016.

\bibitem{zheng2015scalable}
L.~Zheng, L.~Shen, L.~Tian, S.~Wang, J.~Wang, and Q.~Tian.
\newblock Scalable person re-identification: A benchmark.
\newblock In {\em Proceedings of the IEEE international conference on computer vision}, pages 1116--1124, 2015.

\bibitem{Market1501}
L.~Zheng, L.~Shen, L.~Tian, S.~Wang, J.~Wang, and Q.~Tian.
\newblock Scalable person re-identification: A benchmark.
\newblock In {\em Proceedings of the IEEE international conference on computer vision}, pages 1116--1124, 2015.

\bibitem{PRW}
L.~Zheng, H.~Zhang, S.~Sun, M.~Chandraker, and Q.~Tian.
\newblock Person re-identification in the wild.
\newblock {\em arXiv preprint arXiv:1604.02531}, 2016.

\bibitem{RPIfield}
M.~Zheng, S.~Karanam, and R.~J. Radke.
\newblock Rpifield: A new dataset for temporally evaluating person re-identification.
\newblock In {\em 2018 IEEE/CVF Conference on Computer Vision and Pattern Recognition Workshops (CVPRW)}, pages 1974--19742, 2018.

\bibitem{zheng2015partial}
W.-S. Zheng, X.~Li, T.~Xiang, S.~Liao, J.~Lai, and S.~Gong.
\newblock Partial person re-identification.
\newblock In {\em Proceedings of the IEEE international conference on computer vision}, pages 4678--4686, 2015.

\bibitem{PartialREID}
W.-S. Zheng, X.~Li, T.~Xiang, S.~Liao, J.~Lai, and S.~Gong.
\newblock Partial person re-identification.
\newblock In {\em Proceedings of the IEEE international conference on computer vision}, pages 4678--4686, 2015.

\end{thebibliography}
}

\end{document}